\documentclass[1p]{elsarticle}
\usepackage[utf8]{inputenc}
\usepackage[T1]{fontenc}
\usepackage{caption}
\usepackage{subcaption}
\usepackage{float}
\usepackage{verbatim}
\usepackage[section]{placeins}
\usepackage{amsmath}
\title{Targeted Deep Regression - Initial Applications in Whole-Body Water-Fat MRI in UK Biobank}

\author[1,2]{Filip Marttala\corref{cor1}}
\author[3]{Taro Langner}    
\author[1]{Johan Öfverstedt}    
\author[1]{Nouman Ahmad}    
\author[1,3]{Håkan Ahlström}
\author[2]{Robin Strand}
\author[1,3]{Joel Kullberg}

\affiliation[1]{organization={Radiology, Department of Surgical Sciences, Uppsala University},
                postcode={752 37},
                city={Uppsala},
                country={Sweden}}
\affiliation[2]{organization={Department of Information Technology, Uppsala University},
                postcode={751 85},
                city={Uppsala},
                country={Sweden}}
\affiliation[3]{organization={Antaros Medical AB, BioVenture Hub},
                postcode={431 53},
                city={Mölndal},
                country={Sweden}}
\cortext[cor1]{Corresponding author, filip.marttala@it.uu.se}

\begin{document}

\begin{abstract}
UK Biobank is a British clinical study which, in addition to large amounts of medical data collected, plans to include detailed Magnetic Resonance Imaging (MRI) examinations from 100,000 participants. 
The image data can be analyzed for example by automated analysis with neural networks for image segmentation or classification to expand the collection of image-derived phenotypes for medical research. In this work, we propose, present and evaluate a methodology that uses image-based regression to estimate numerical properties from the collected neck-to-knee water-fat MRI scans. 
It applies a ResNet50-style network to 2D multi-channel collages that represent targeted and study-specific anatomical sub-regions from the MRI scan. 
The targeted sub-region sampling is achieved by propagating manually defined subregions from a reference scan to the other >30,000 analyzed scans using image registration. 
The resulting collages can be compact while retaining high-resolution, for example by providing multiple slices of kidney tissue for a kidney volume regression network. 
This approach also enables group-wise saliency analysis that highlight image and anatomical regions used by the network for the predictions. 
The method was applied for the regression endpoints of kidney volume, heart volume, and age. Predictive performance was found between $R^2=0.731 \text{---}0.984$ and the group-wise saliency maps were found to highlight medically relevant regions. We conclude that the method can be used to standardize and restrict the information presented to a deep regression analysis pipeline, increasing information density and opening up a wide range of future medical inference studies.

\end{abstract}

\flushbottom
\maketitle
%
%
\thispagestyle{empty}

\newpage

\section{Introduction}

A popular method of medical imaging is Magnetic Resonance Imaging (MRI) which uses magnetic fields and radio-waves to generate images with good soft tissue contrasts. Its non-invasive nature and the fact that it does not expose the scanned subjects to ionizing radiation makes it very useful both in clinical practice and research. Because of this, it is possible to conduct large-scale imaging studies on healthy individuals, such as UK Biobank \cite{sudlow2015uk, littlejohns2020uk}. 
Further, MRI can produce 3D images with detailed anatomical information about organs, vessels and bone at high resolution. While large-scale collection of images is possible, however, large-scale analysis of said images is challenging. Manual analysis of MRI can require minutes or hours per image, making it slow or limited in scope to relatively small sample sizes. 

Machine learning techniques can provide major benefits by automating some of the required processing steps for image analysis. In the UK Biobank, proposed use cases for automation via neural networks include semantic segmentation for organ measurements \cite{bai2018automated, langner2020kidney, liu2021genetic, kart2021deep} and quality control \cite{basty2020image}, but also image-based, deep regression \cite{langner2022mimir}. The latter can directly estimate numerical endpoints from image data, enabling a range of different predictions \cite{9090311,cole2018brain}. Whereas segmentation networks can perform pixel- or voxel-wise predictions, image-wise predictions by regression networks typically require larger quantities of training data \cite{langner2022mimir}. This is especially the case for volumetric MR images with millions of voxel values, each of which could affect the prediction. Although image-based deep regression has been previously proposed for volumetric MRI \cite{armanious2021organ, nimhed2022estimation}, images are often compressed or restrict to 2D.

One major downside of neural networks is that it is often not clear what features are relevant for a given prediction \cite{selvaraju2017grad}. This may be considered problematic since a network prediction does not inherently provide any insights into the logic behind this inference. This may reduce confidence in the prediction as well as making it more difficult to catch faulty predictions. This problem can be partially solved by the use of saliency maps. Saliency maps are a method of highlighting image regions with high impact on the network prediction, typically based on backpropagation of gradients in the network output back to the pixel inputs \cite{selvaraju2017grad}. This can result in a map highlighting which input pixels had a large impact on the prediction. Individual saliency maps might however be noisy and difficult to interpret. If a large number of saliency maps are averaged, however, it is possible to pick out general patterns of general relevant features of the images\cite{langner2019identifying}.

The main contribution of this work consists in combining image-based, deep regression and saliency analysis with a new type of input image format that represents volumetric MRI data in 2D collages showing anatomy that is targeted and specific to the given regression task. The method allows for manually defining specific regions of interest from a UK Biobank neck-to-knee MRI and automatically extracting the anatomically corresponding region in other images by leveraging image registration\cite{ekstrom2020fast}. These regions are then used to generate a 2D multi-channel image and used as input for a ResNet50-style neural network. The approach was trained and evaluated on several endpoints for more than 30,000 UK Biobank neck-to-knee body MR images.

\newpage
\section{Methods}

\subsection{Overview of the targeted deep regression approach}

\begin{figure}[H]
    \centering
    \includegraphics[width=\textwidth]{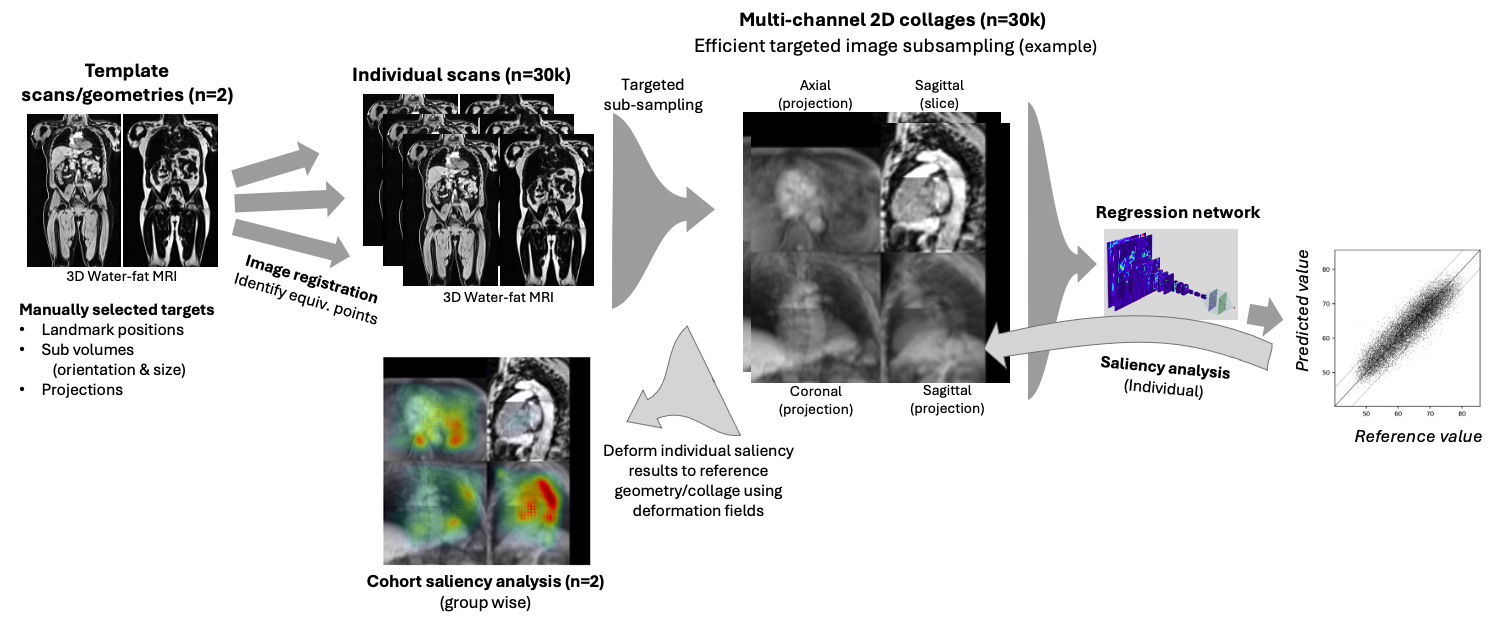}
    \caption{Illustration of the image analysis pipeline used for the targeted deep regression, with associated cohort saliency analysis that enables interpretability. Template scans (one male and one female) from the 3D water-fat MRI images in UK biobank are first selected. The targeted subsampling of the image data is manually defined using these. Image landmarks, subvolumes and projection are selected and used to target specific anatomical regions of the UK Biobank images and to convert the 3D image data into a 2D representation (e.g. axial mean intensity projection of a subvolume centered on the heart). Deformable image registration is used to transfer the manually selected landmarks and subvolumes into all individual scans. This allows targeted subsampling of the 3D data into a 2D representation from all scans. The extracted multi-channel 2D images are combined into a multi-channel collage for every scan. These collages are used as input to a regression network that is trained to predict a reference value of interest from each individual collage (e.g. the age of a subject). The trained regression networks are furthermore analyzed by performing saliency analysis for each performed inference. This gives an individual saliency result, i.e. a heatmap, for each collage, of what regions in a collage important for the performed task. By use of the image registration results these individual saliency results can be combined and aggregated into a cohort saliency map that shows a cohort-wise result of what regions that are important for the predictions (e.g. what regions are found most important for the age prediction task). }
    \label{fig:samplingexplanation}
\end{figure}

An overview of the targeted deep regression approach is given in Figure \ref{fig:samplingexplanation}. By use of image registrations manually selected target image slices and volumes is subsampled for all individuals and scans on a cohort study. The targeted image data is combined into a multi-channel 2D collage that is used as input to a regression network. Saliency analysis is used for interpretability. The image registrations furthermore enable group-wise or cohort saliency analysis.

\newpage
\subsection{Data sets}

The MRI data and clinical data used in this study originates from UK Biobank, a large-scale study that collects health information on over half a million volunteers aged 44-82 years \cite{sudlow2015uk}. For 100,000 of these participants, further data collection is planned, including, but not limited to, whole-body MRI scans with neck-to-knee coverage. This study included over 40,000 of these whole-body MRI scans. The images were collected using a dual-echo Dixon protocol with a 1.5 T MR Siemens Aera device. These scans allow analysis of images of water signal, fat signal as well as quantitative water fraction as well as fat fraction images\cite{west2016feasibility}, with
$\text{TR} = 6.69 \text{ms}$, 
$\text{TE} = 2.39\text{/}4.77 \text{ms}$ and
$\text{flip angle} = 10$°. 
The dimensions of each voxel are 2.23mm $\times$ 2.23 mm with the last dimension varying, being 3 mm, 3.5 mm, or 4.5 mm \cite{littlejohns2020uk}. For the purposes of having consistent sizes, all images were resampled to an effective voxel resolution of 2.23mm $\times$ 2.23mm $\times$ 3mm (SAG $\times$ COR $\times$ AX).

The numerical values that served as ground truth for the regression targets of age, kidney volume, and heart volume were gathered from several different sources. For age, the UK Biobank metadata provided only the year and month of birth (UK Biobank fields 34 and 52). Every subject was therefore assumed to be born in the middle of the month, so that any error in the ground truth data amounts to at most half a month \cite{langner2019identifying}. The values for kidney volume were based on prior work that utilized convolutional neural network for segmentation of the same neck-to-knee body MRI as used here \cite{langner2020kidney}. The measurements of the heart consist of the left ventricular end-diastolic volume (LV-EDV) (UK Biobank field 22421) \cite{suinesiaputra2018fully}. It is worth noting that the ground truth dataset for LV-EDV is noisy and UK Biobank includes a warning of non-quality controlled data for this parameter, with the original set of reference measurements containing values from a few 10s of mL to 15,000 mL (see UK Biobank field 22421). To limit the impact of these errors, the data was therefore filtered to exclude any participant with a value higher than 260 ml. 

The UK Biobank study and the experiments performed here were approved by the British and Swedish ethics committees, respectively. The subject gave written informed consent, and the study has been performed according to the Declaration of Helsinki\cite{sudlow2015uk}. The UK Biobank data was accessed via the UK Biobank website\footnote{https://www.ukbiobank.ac.uk/} and their associated Access Management System\footnote{https://ams.ukbiobank.ac.uk/ams/}. Data is de-identified/pseudo-anonymized by the UK Biobank organization before it is shared.

\begin{figure}[ht]\label{fig:collageexamples}
\centering
  \begin{subfigure}[b]{0.3\textwidth}
    \includegraphics[width=\textwidth]{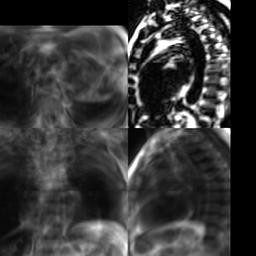}
    \caption{Fat image}
    \label{fig:heartFat}
  \end{subfigure}
 \hfill
  \begin{subfigure}[b]{0.3\textwidth}
    \includegraphics[width=\textwidth]{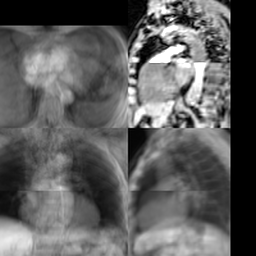}
    \caption{Water image}
    \label{fig:heartWater}
  \end{subfigure}
 \hfill
  \begin{subfigure}[b]{0.3\textwidth}
    \includegraphics[width=\textwidth]{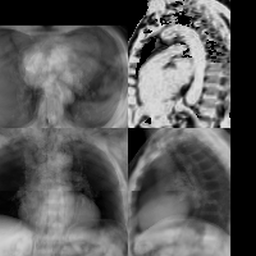}
    \caption{Water Fraction image}
    \label{fig:heartFraction}
  \end{subfigure}  
  \caption{An example of a heart collage illustrating the three image channels used (a, b, and c). In this example three projections are used in each channel. Axial (top left), coronal (bottom left), sagittal (bottom right) as well as one sagittal slice (top right) defined using a landmark in the (Aorta).}
  \label{fig:heartcollage}
\end{figure}

\newpage
\subsection{Targeted image collages}

The input images provided to the neural network were formatted as targeted collages, i.e. the format was designed specifically for a given regression task, based on a research hypothesis. This was done by manually targeting specific regions of interest to be sampled, such that any anatomical area important for the measurement was in the targeted volumes, see Figure \ref{fig:samplingexplanation}. A rectangular region of interest was defined and specified by its midpoint and side lengths in the male and female reference subjects respectively. Similar samples were then taken from the remaining images by using deformation fields to map the midpoint of the region of interest to its respective location in the non-reference images. For each image, a region of interest of the same size was then sampled around this new midpoint.
Because the network requires the input to be a 2D image, sampled 3D regions were then projected by calculating the mean intensity value along a specified direction. 

\subsection{Image registration}

Image registration was used to map manually defined regions of interest from a reference image to corresponding regions in all other images. 
The image registrations were created by selecting a reference image/subject from each sex and registering all other images to the reference image of the corresponding sex. 
The reference images were chosen to have an average tissue composition and were quality controlled for artifacts. 
The image registration used a deformable image registration method based on graph-cuts utilizing the software \textit{deform}\cite{ekstrom2020fast} that generated, for each image, deformation field mappings (vector fields) between a sex-specific reference and a given target image.
Each deformation field describes how points in the reference images map to the target image and can be used to map manually defined coordinates in the reference persons to corresponding points in all other images.




\subsection{Multi-channel collage construction}

After sampling all desired regions and projecting them onto 2D, the resulting 2D images of targeted anatomical areas were arranged into one collage for each MR image, as shown in Figure 2. These collages were fed to the neural network.
Minimizing the size of the collage is important for performance reasons and is a packing problem. This minimization was done using the rectangle-packer package for Python\footnote{https://pypi.org/project/rectangle-packer/}. When a satisfactory packing was been achieved the water signal, fat signal as well as either water fraction or fat fraction images are combined into a 3 channel 256 $\times$ 256 image. 
For storage reasons the images were saved in 8-bit unsigned integer format. This was not found to have any noticeable negative impact on the predictive performance of the network. Because the intensity values range from $0$ to about $1500$, the intensity values were scaled down such as to fit in the $0$-$255$ range of the uint8 format. For the fat and water image this was done by normalizing the image intensities after clipping the top 1\% of the brightest voxel values and any negative values potentially resulting from the spline interpolation. 

\subsection{Network}

A ResNet50 \cite{he2016deep} architecture was used to perform the regression with pre-training being done on ImageNet \cite{langner2022mimir}. The network was trained using the Adam optimizer \cite{kingma2014adam} for 10,000 iterations with the learning rate decreasing by a factor 10 after 8,000 iterations. The training code is publicly available\footnote{github.com/tarolangner/ukb\_mimir}. We used a batch size of 20 due to the limited memory of the NVidia 1060 GPU which has 6GB of memory. The time to train the network for 10,000 iterations was approximately 1.5 hours.

\subsection{Evaluation}


The evaluation was performed using 10-fold cross validation\cite{Refaeilzadeh2016}, for a setup of training on 90\% of the data and validating on the remaining 10\% in a rotating scheme. To evaluate the predictive performance of the network we used the mean absolute error (MAE) and mean absolute percentage error (MAPE) to measure the average accuracy of the predictions. The latter was defined as the mean over all absolute differences between prediction and ground truth, each divided by the corresponding ground truth. Pearson's coefficient of correlation (r), the coefficient of determination ($R^2$), and the 95\% limits of agreement (LoA) were used to compare the model predictions and the known ground truth values. Bland-Altman plots were also generated to evaluate the results. 

\subsection{Group-wise saliency analysis}   

The saliency analysis was performed for each individual regression task by simple backpropagation using built-in methods in PyTorch. The individual saliency maps were combined from all inferences in the cross-validation into group-wise average saliencies for males and females, respectively. For each image in each individual collage, the saliency map was deformed back using the performed image registration to standardize the geometry. This was done by first mapping each image back to the reference by inverting the deformation fields using simpleitk \cite{lowekamp2013design, yaniv2018simpleitk, beare2018image}. This inversion was performed on each sample after projection, thus requiring the inversion of only a 2D field, greatly reducing the computational cost. This also allows us to create a deformation field collage mapped the same as the image collage.



\newpage
\subsection{Experiments}

To evaluate the feasibility of the method a set of experiments, increasing order of complexity, was performed. The experiments were:

\begin{itemize}
    \item A - Kidney volume prediction using a collage of the kidneys
    \item B - Left Ventricular End Diastolic Volume (LV-EDV) prediction using a collage of the heart
    and surrounding regions
    \item C - Age prediction using a collage of sagittal slices of the heart and surrounding regions
    \item D - Age prediction using a collage of axial slices of the heart and surrounding regions
\end{itemize}

The first two experiments were performed with clear prior knowledge of what results to expect, as the
desired information was known to be available in the image data used, even if some of it may have been lost in the projection process. The last two were of more exploratory nature to test the limits of the proposed approach. Here, different orientations of the image sampling were used to demonstrate that the experiment design allows for different structures to be sampled, such as the aorta diameter as well as its length and curvature, for example. 

The collages used in Experiments A-D are shown in figure 3.

\begin{figure}[!htb]
    \centering
    \includegraphics[width=\textwidth]{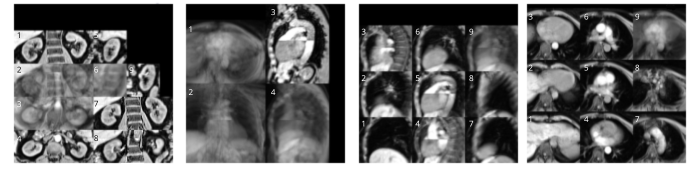}
    \caption{Examples of water channels of the collages used in the experiments, for kidney volume (far left), left ventricular end diastolic volume (LV-EDV) of the heart (center left), and age from images targeted on the heart using sagittal slices (center right) and axial slices (far right). Note that the packing algorithm used results in a collage with the different image in no particular order. The numbers are not part of the collage image data. They are added to simplify description of the different sub images and referred to as (x) in the text.}
    \label{fig:combined saliency}
\end{figure}

\newpage
\subsection*{Detailed description of experiments}

\subsection*{A - Kidney volume prediction}
The kidney collage is sampled from a subvolume of size $269.83\times 113.73\times 153.00$mm (SAG $\times$ COR $\times$ AX), centered between the kidneys. The targeted sampling consisted of three coronal slices, offset by $10.6$mm $(1),(7),(8)$, an axial slice $(4)$, two sagittal slices offset to capture the center of each kidney $(5)$ and $(9)$. The collage also contains mean projections in the coronal $(2)$, axial $(3)$ and sagittal $(6)$ directions, respectively.


\subsection*{B - Left Ventricular End Diastolic Volume (LV-EDV) prediction}
The heart collage is sampled from a subvolume of size $225.23\times 180.63\times 303$mm and is made up by a coronal mean projection $(1)$, an axial mean projection $(2)$, a sagittal mean projection $(4)$ as well as a slice in the sagittal direction $(3)$ aimed to try to sample the bent aorta.


\subsection*{C - Age prediction from sagittal slabs}
The sagittal heart slab collage consisted of eight contiguous slabs of thickness 22.3mm that all together span a volume of $180.63\times 167.25\times 195$mm centered around the heart. Subsamples 1 through 8 span from right to left. Subsample 9 is a mean projection of the sample volume in the sagittal direction. One voxel from the volume is missing from the slabs in favor of making all slabs the same thickness. 


\subsection*{D - Age prediction from axial slabs}
The axial heart slab collage consisted of eight contiguous slabs of thickness 24mm that altogether span a volume of $171.71\times 167.25\times 195$mm centered around the heart. Subsamples 1 through 8 span from inferior to superior. Subsample 9 is a mean projection of the sample volume in the axial direction. One voxel from the volume is missing from the slabs in favor of making all slabs the same thickness. 

\begin{figure}[H]
    \centering
    \includegraphics[width=1.0\linewidth]{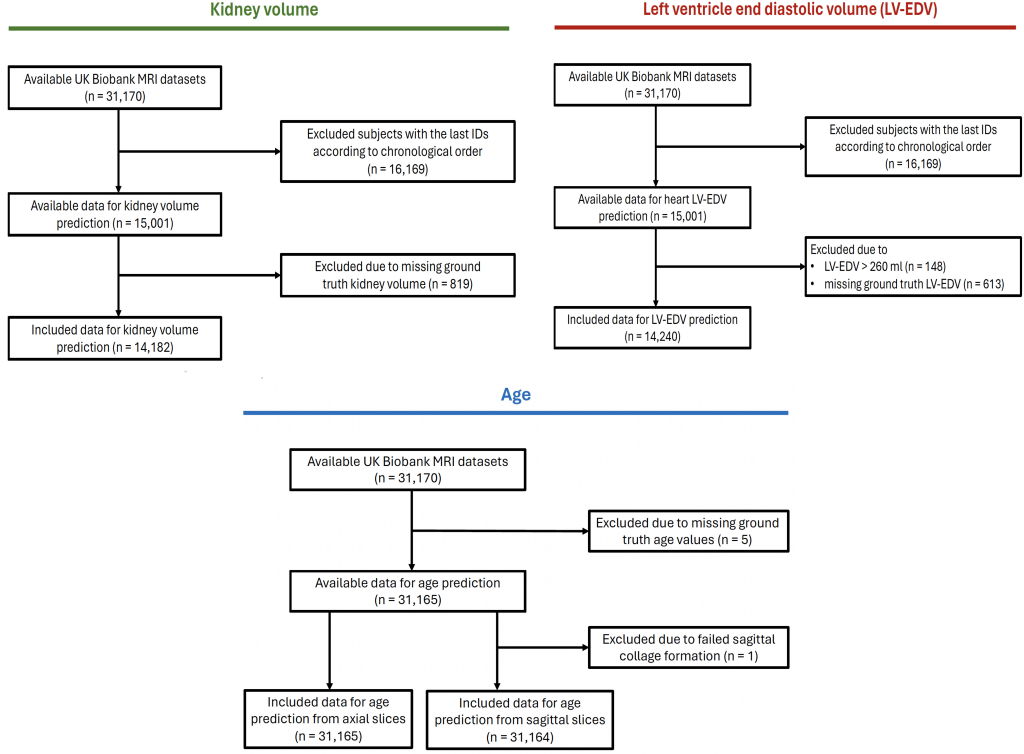}
    \caption{STROBE flowchart showing how the participants for experiments were selected}
    \label{fig:stobe}
\end{figure}

The selection of participants are illustrated in Figure \ref{fig:stobe}.


\section{Results}

Evaluation metrics for all regression targets are summarized in Table \ref{tab:evmetrics}. Input collages were furthermore visualized for each of the four experiments, overlaid with the corresponding aggregated saliency maps, together with scatter plots and Bland-Altman plots in the Figures \ref{fig:kidneyvolume}, \ref{fig:heartlvedv}, \ref{fig:heartagesagittal}, and \ref{fig:heartageaxial}, respectively.

For the kidney volume experiment the results were highly accurate, with a mean absolute error of 6.8 mL and a mean absolute percentage error of about 2.8\%. Further, the saliency maps showed that the network was able to locate and use the image information in kidneys with rather high accuracy. 

The prediction of LV-EDV was not as accurate as for kidney volume, with a mean absolute error of 12.3 mL and a mean absolute percentage error of 9.8\%. The saliency maps show that the method is able to localize and use information from region of the left ventricle.

Regarding the age predictions they resulted in similar performance using sagittal and axial slicing, with the latter performing slightly numerically superior. The group-wise saliency maps were also seen to highlight similar regions using both slice orientations. The regions were identified as left atrium, spinal vertebras and a region in the lungs was found highlighted in both males and females using both slice orientations.



\begin{table}[ht]
\centering
\begin{center}
\begin{tabular}{lcc|rcccc}
Target                   & N  & Unit & MAE      & MAPE          & r     & $R^2$  & LoA            \\ 
\hline
Kidney volume            & 14,182   &  mL & 6.761     &    0.0277      & 0.984 & 0.967  & (-18.84 to 19.37) \\
Heart LV-EDV             & 14,240   &  mL & 12.324      &   0.0981    & 0.856 & 0.731  & (-32.41 to 33.15) \\
Age sagittal slices& 31,164   &  years & 2.832     &     0.0448     & 0.880  & 0.773  & (-6.99 to 7.06)   \\
Age axial slices   & 31,165    &  years & 2.759     &    0.0437      & 0.887 & 0.790 & (-6.92 to 6.72)
\end{tabular}
\end{center}
\caption{Summary of evaluation metrics from all four experiments, listing number of included subjects (N), mean absolute error (MAE), mean absolute percentage error (MAPE), Pearson's coefficient of correlation (r), the coefficient of determination ($R^2$) and the 95\% Limits of Agreement (LoA) from the 10-fold cross-validation.}
\label{tab:evmetrics}
\end{table}

\newcommand{\resultcollages}[7]{
    \centering
    \begin{subfigure}[b]{0.24\textwidth}
        \includegraphics[width=\textwidth]{#1}
        \caption{Male collage and saliency}
        \label{#2}
    \end{subfigure}
    \begin{subfigure}[b]{0.24\textwidth}
        \includegraphics[width=\textwidth]{#3}
        \caption{Female collage and saliency}
        \label{#4}
    \end{subfigure}
    \hfill
    \begin{subfigure}[b]{0.43\textwidth}
        \includegraphics[width=\textwidth]{#5}
        \caption{Scatter plot (left) and Bland-Altman plot (right)}
        \label{#6}   
    \end{subfigure}
    \caption{#7 prediction results for males \ref{#2} and females \ref{#4}, with plotted numerical results \ref{#6}.}
} 
\newpage
\begin{figure}[H]
    \resultcollages{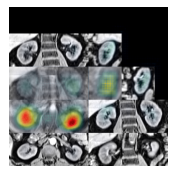}
    {fig:kidneyvolumesaliencyM}
    {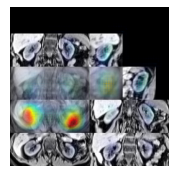}
    {fig:kidneyvolumesaliencyF}
    {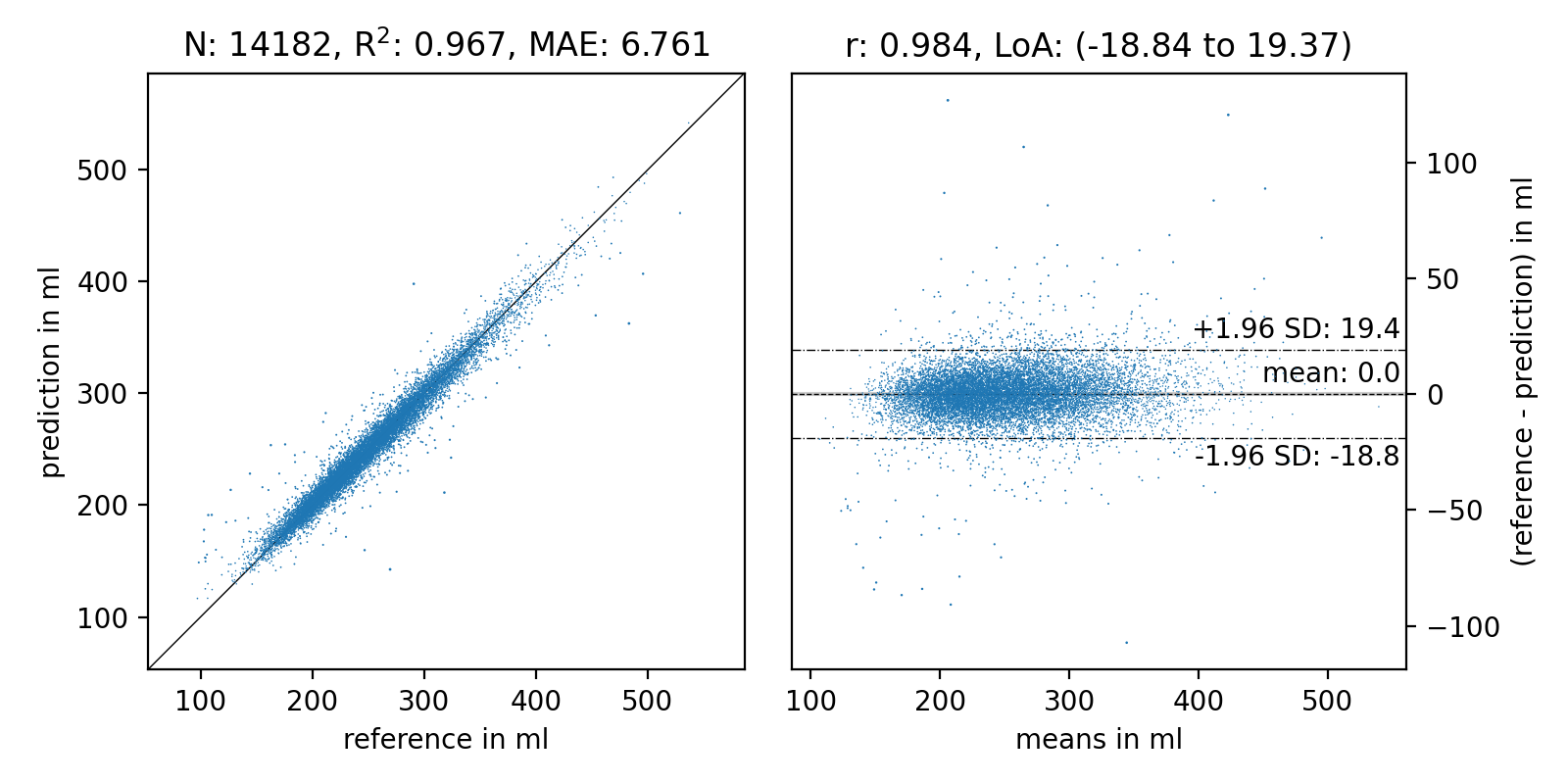}
    {fig:kidneyvolumeresults}
    {Kidney volume}
    \label{fig:kidneyvolume}
\end{figure}

\vspace{-0.8cm}

\begin{figure}[H]
    \resultcollages{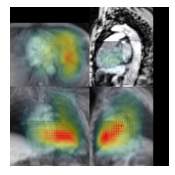}
    {fig:HeartLvedvM}
    {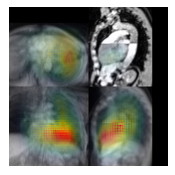}
    {fig:HeartLvedvF}
    {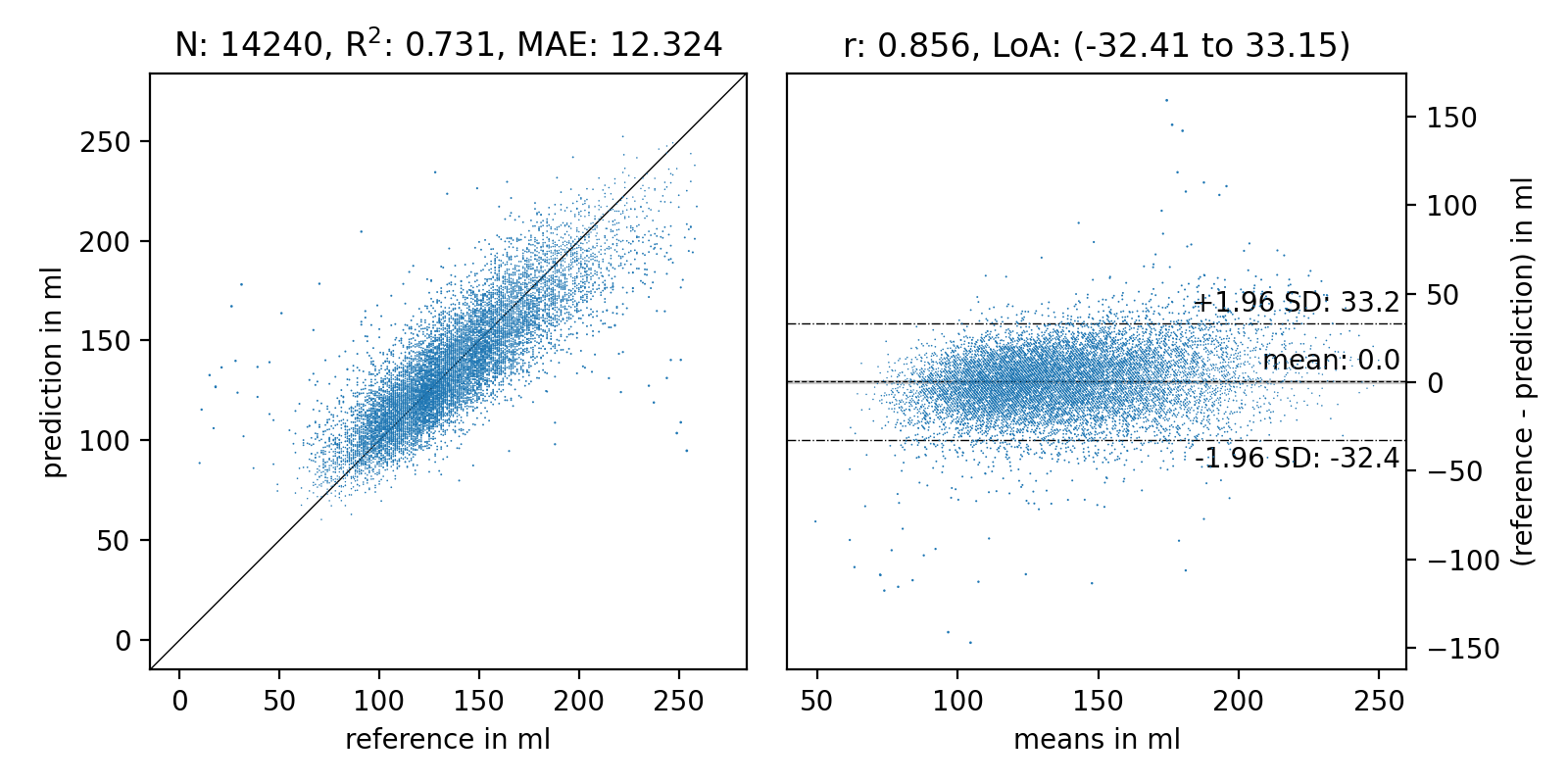}
    {fig:heartlvedvresults}
    {Heart LV-EDV}
    \label{fig:heartlvedv}
\end{figure}

\vspace{-0.8cm}

\begin{figure}[H]
    \resultcollages{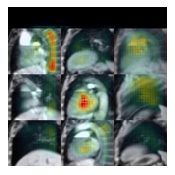}
    {fig:SagittalHeartAgeM}
    {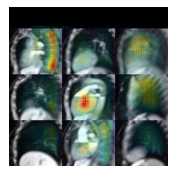}
    {fig:SagittalHeartAgeF}
    {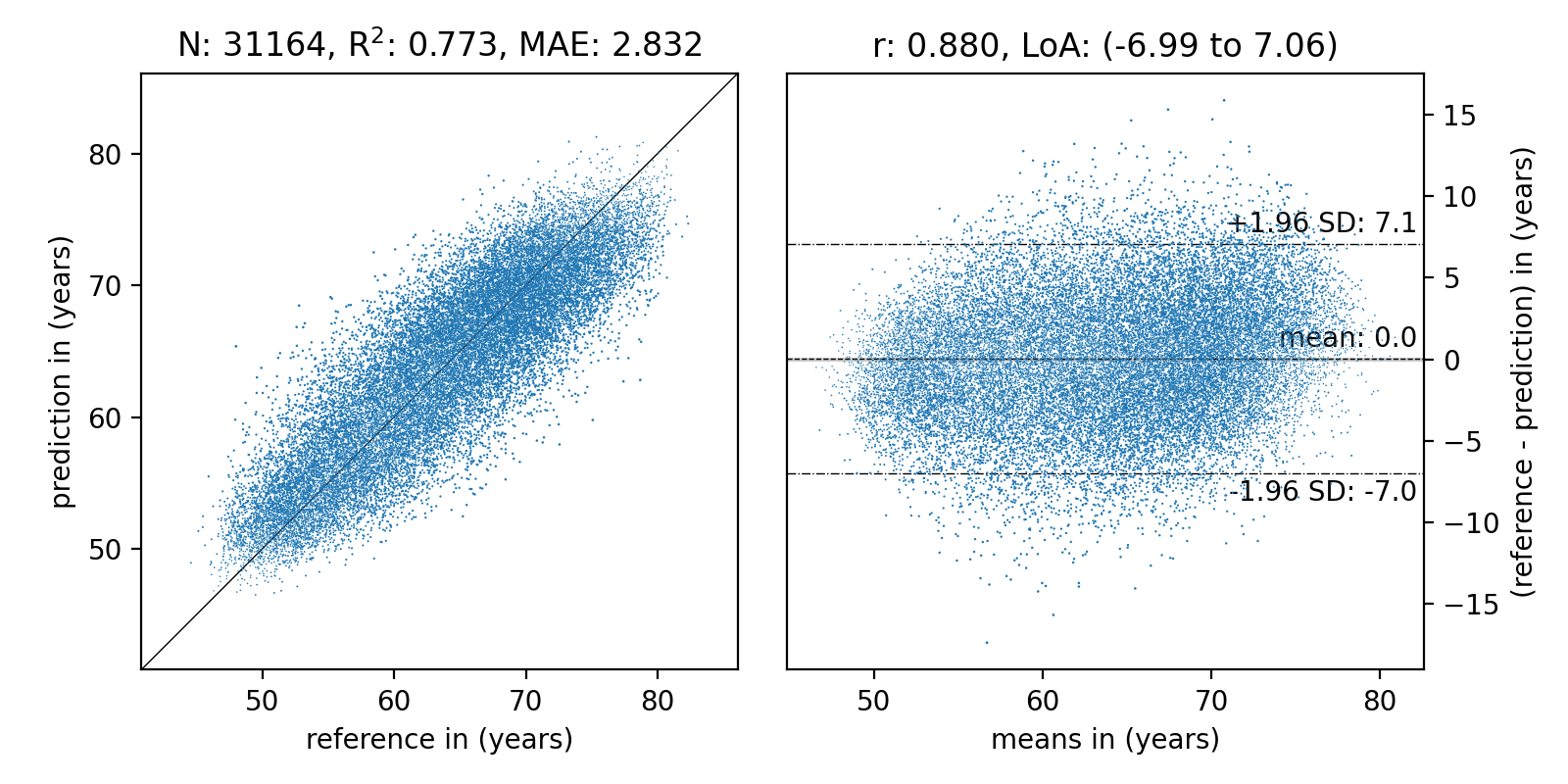}
    {fig:heartagesagittalresults}
    {Age (sagittal)}
    \label{fig:heartagesagittal}
\end{figure}

\vspace{-0.8cm}

\begin{figure}[H]
    \resultcollages{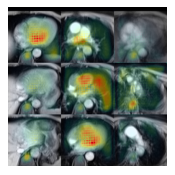}
    {fig:AxialHeartAgeM}
    {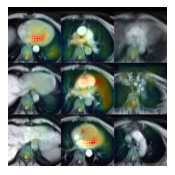}
    {fig:AxialHeartAgeF}
    {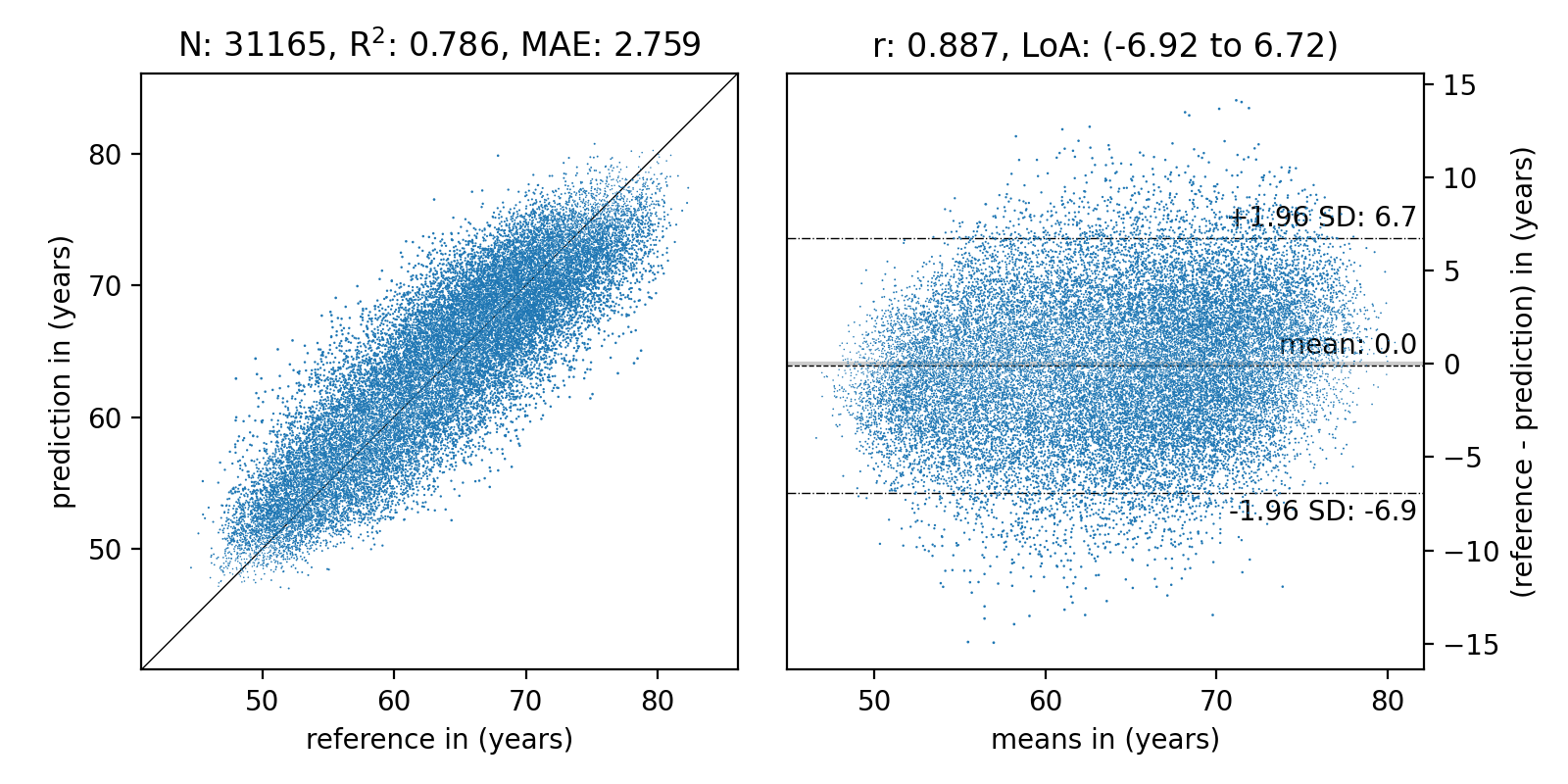}
    {fig:heartageaxialresults}
    {Age (axial)}
    \label{fig:heartageaxial}
\end{figure}
\newpage

\section{Discussion}

We have in this work presented and evaluated a methodology for targeted deep regression using tailored collages that can represent specific morphological information of MR image data and highlight anatomical structures relevant for the prediction with saliency analysis. The proposed technique formats the MRI data into one 2D collage for each subject, based on manually selected regions of interest that are defined in one male and one female template subject and then propagated to all other neck-to-knee body MRI scans with the help of image registration. Predictive performance was found between $R^2=0.731 \text{---}0.984$ and the group-wise saliency maps were found to highlight medically relevant regions.

The kidney volume regression achieved high accuracy and distinctly highlighted salient kidney tissue in the axial mean intensity projection of the collages, as seen in Fig. \ref{fig:kidneyvolumesaliencyM} and Fig. \ref{fig:kidneyvolumesaliencyF}. The mean absolute percentage error of 2.8\% achieved here rivals the intra-operator variability achieved by human annotators cited in the original work that provided the underlying kidney volume measurements based on neural networks for semantic segmentation \cite{langner2020kidney}. It also clearly outperformed a previously proposed approach that reported a relative error of 7\% for the same kidney volume measurements, based on a more generic deep regression approach on mean intensity projections of the UK Biobank neck-to-knee body MRI \cite{langner2022mimir}. It is possible that the projection images are on average more useful for a network to focus on as it shows less variation in the presentation compared to a sampled slice determined using image registration as a projection image contains both depth and width information in intensity and in-slice span.

The prediction accuracy of LV-EDV was somewhat lower than when predicting kidney volume. This was expected since LV-EDV is more challenging due to the nature of the left ventricle volume being a subregion of the heart as well as not being constant over the heart beat. This means the network must not only be able to recognize the heart but also its smaller sub-volumes. Further, the UK Biobank reference values were originally acquired using dedicated cardiac MRI scans collected using ECG-triggered scanning \cite{suinesiaputra2018fully}. This means that the images used here for predictions are less suitable than the ones used for collecting the ground truth, leading to worse results. As we seen in the saliency maps of Fig. \ref{fig:HeartLvedvM} and \ref{fig:HeartLvedvF}, however, the network was on average capable of locating the heart as well as recognizing the left ventricle (located in the lower right part of the heart in the images) in the collages.

The age regression experiments yielded predictions with a mean absolute error just below 3 years, using image data from the mediastinal region only. In related work, a volumetric age regression approach on UK Biobank neck-to-knee body MRI yielded a mean absolute error of 3.6 years \cite{nimhed2022estimation}, whereas earlier work on 2D mean intensity projections achieved about 2.5 years \cite{langner2019identifying}. The latter found aggregated saliency analysis highlighting predominantly structures of the aortic arch and knees relevant for age. 

In the experiments performed in the current study we note high saliency in the heart (left atrium) in spine and lung tissue, with little emphasis on the aortic arch. These patterns were seen in both females and males and in the experiment using axial and sagittal projections (see Fig. \ref{fig:SagittalHeartAgeM} and Fig. \ref{fig:SagittalHeartAgeF}).

The targeted sampling can maximize the amount of relevant information in the created collage by targeting specific regions, here kidney and heart for the volume estimation experiments. It can also tailor an experiment for machine-learning based studies of associations between certain organs or tissues and specific biomarkers of interest, such as the thoracic region and age examined here. For both the kidney and heart (LV-EDV) volume experiments the relevant information was readily available in the source image data, limited only by the image quality, e.g. the non-ECG-triggered scans of the whole-body volume. The created collages also aimed to maximize the information in the input format for the neural network, though with the risk that the projections used would result in too much information being lost. Furthermore, these experiments were also made with strong prior knowledge on where the network could be expected to focus its attention to optimize the predictions.

We note that the in the volume predictions illustrated in Figures 4 and 5 the groups-wise saliency tends to prefer projections over slices. We hypothesize that this might be due to the fact that the projections contain information from larger volumes compared to slices and thus contain more useful information for volume predictions. It might also be that individual slices show more variability in content and appearance that makes it more difficult for a network to use in the predictions. This illustrates the potential of using different targeting or subsampling strategies to compare the "importance" of different regions and sampling strategies for the predictions.

One limitation of the image formatting used here consists in the need for two reference images, one for each sex. Due to this, two targeted samplings are necessary which cannot be strictly guaranteed to be identical. This might limit studies on, for example, sex-related differences. The sex-specific sampling may lead to the sex being obvious even without learning any anatomical differences between the sexes. If, as an extreme case, the heart of all male subjects was sampled but the corresponding region of interest for female subjects was placed with a significant offset, the network may easily predict sex without actually learning anything of use for the intended study. Future work could focus on mitigating this problem, for example by registering one of the references to the other. Then only one image would need to be sampled since the sampling could be automatically transferred to the other reference, similarly to how it is done with the rest of the data set.

In conclusion we have presented and evaluated a methodology for tailored deep regression that can be used to specify and restrict the image data presented to a 2D deep regression analysis pipeline, allowing higher information density and a wide range of future medical inference studies. Beyond just estimating missing measurements, the presented saliency mapping technique provides a visualization of the "black box" analysis performed. The resulting saliency maps enable a biological interpretation and hypothesis generation, opening up a wide range of potential future regression experiments that can integrate imaging and non-imaging information.



\newpage

\bibliography{sample.bib}

\section*{Acknowledgements}

This research has been conducted using the UK Biobank Resource under Application Number 14237 and the Swedish National Infrastructure for Computing (SNIC, sens2019016). This research was enabled by grants from EXODIAB, the Swedish Heart-Lung Foundation (2020-0492) and the Swedish Research Council (2016-01040, 2019-04756, 2023-03607 ).

\section*{Author contributions statement}

F.M. wrote the main manuscript text and conducted the experiments, N.A. contributed with methodology and code, J.Ö. provided methodological expertise and edited the manuscript, T.L. developed the previous version of the methodology and edited the manuscript, H.A. provided medical expertise., R.S. and J.K. initiated and supervised the project and the data access from UK Biobank. All authors read and approved the final manuscript.

\section*{Competing interests}

JK, HA, and TL are or have been employed by Antaros Medical during the project period. Antaros Medical owns the right to patents related to the image registrations used in this work. FM, NA and JÖ report no competing interests.

\end{document}